# Myocardial Infarction Quantification From Late Gadolinium Enhancement MRI Using Top-hat Transforms and Neural Networks

Ezequiel de la Rosa, Désiré Sidibé, Thomas Decourselle, Thibault Leclercq, Alexandre Cochet, Alain Lalande*

*Abstract*— *Significance:* **Late gadolinium enhanced magnetic resonance imaging (LGE-MRI) is the gold standard technique for myocardial viability assessment. Although the technique accurately reflects the damaged tissue, there is no clinical standard for quantifying myocardial infarction (MI), demanding most algorithms to be expert dependent.** *Objectives and Methods:* **In this work a new automatic method for MI quantification from LGE-MRI is proposed. Our novel segmentation approach is devised for accurately detecting not only hyper-enhanced lesions, but also microvascular-obstructed areas. Moreover, it includes a myocardial disease detection step which extends the algorithm for working under healthy scans. The method is based on a cascade approach where firstly, diseased slices are identified by a convolutional neural network (CNN). Secondly, by means of morphological operations a fast coarse scar segmentation is obtained. Thirdly, the segmentation is refined by a boundary-voxel reclassification strategy using an ensemble of CNN's. For its validation, reproducibility and further comparison against other methods, we tested the method on a big multi-field expert annotated LGE-MRI database including healthy and diseased cases.** *Results and Conclusion:* **In an exhaustive comparison against nine reference algorithms, the proposal achieved state-of-the-art segmentation performances and showed to be the only method agreeing in volumetric scar quantification with the expert delineations. Moreover, the method was able to reproduce the intra- and inter-observer variability ranges. It is concluded that the method could suitably be transferred to clinical scenarios.**

## I. INTRODUCTION

Late gadolinium enhancement (LGE) MRI is the cornerstone of myocardial tissue characterization [1], representing the most accurate and highest resolution method for myocardial infarction (MI) and non-ischemic cardiomyopathies diagnosis. It allows, as well, risk stratification and outcome prediction after revascularization processes or cardiac resynchronization therapy. Currently,

LGE inversion recovery and phase sensitive inversion recovery are considered as the gold standards for myocardial viability assessment [2]. Imaging is generally conducted after 10 minutes of gadolinium injection which over-enhances infarcted myocardium by accumulation of the agent in the damaged tissue. In healthy tissue areas, given the fast gadolinium wash in and wash out no agent abnormal accumulation is presented and the normal myocardium remains with hypointense signaling [3]. By means of experimental studies, it was exhibited that the contrast distribution accurately reflects pathology of the myocardium [4]. Infarcted tissue may also present hypointense regions as a consequence of the permanent microvascular obstruction (MVO, also called no-reflow) phenomenon. MVO evidences the lack of reperfusion of some myocardial area even after the ending of the ischemic event, indicating severe ischemic disease and being associated with poor prognosis, adverse cardiac events and remodeling [3], [5].

The main limitations of LGE-MRI for myocardial tissue assessment are not only due to technical parameters setting (such as slice thickness, inversion recovery, etc.) [6] but mainly due to the lack of a clinical standard for scar tissue quantification [2],[6]. Thus, nowadays there is no reference method for abnormal tissue detection and segmentation, even though several techniques have been explored. The most frequently used techniques are the threshold-based ones, such as the full-width at half-maximum (FWHM) [7] and the n-standard deviations (from now on n-SD) [4]. Nevertheless, these methods provide poor agreement with expert delineations, inconsistent and high result variability and significant differences when compared with ground truth [8],[9]. Since most of these methods are manual or semi-automatic requiring visual assessment and human interaction, the lesion quantification process turns tedious, subjective and hardly reproducible. Even more, current proposals are developed for working only over diseased scans where *i)* visual identification of diseased myocardium before algorithm application is required and *ii)* quantification of hyper-enhanced necrotic tissue is only addressed, without taking into consideration the MVO areas.

In this work, a new LGE-MRI infarction quantification method is devised. The main contributions of this work are: *i)* the automatic myocardial lesion detection and quantification

E. de la Rosa was with Le2i, University of Burgundy, France. He is now with Technical University of Munich (Germany) and icometrix (Belgium). D. Sidibé is with Le2i, University of Burgundy, France. T. Decourselle is with Casis company, Dijón, France. T. Leclercq is with the University Hospital of Dijon, France. A. Cochet and *A. Lalande are with Le2i, University of Burgundy, France and with the University Hospital of Dijon, France (correspondence e-mail: alain.lalande@u-bourgogne.fr).

by a robust approach, *ii)* the a-priori discrimination of healthy and diseased myocardium slices, which extends the algorithm application for working under healthy scenarios, *iii)* the incorporation of a dedicated MVO inclusion step, and *iv)* the validation of the algorithm on a large multi-field database accounting with hyper-enhanced and MVO ground truth areas.

## II. BACKGROUND

This section is split into two main fields: *i)* the analysis of myocardial damage detection techniques and *ii)* the current state-of-the-art algorithms for infarction quantification.

### A. Myocardial Damage Detection

Segmentation of the myocardial scar has been widely addressed in many works, where methods have been only validated under pathological datasets. However, before performing the scar segmentation, myocardial abnormality identification by visual inspection is a compulsory step, as in n-SD [4] and FHWM [7] methods. By means of this approach, the lesion's search is guaranteed in abnormal myocardiums only, reducing potential false positives in healthy slices. Despite the fact that these methods are able to control better the false positive rate, one of the drawbacks is the required expert interaction. Under this scenario the development of an automatic method that could deal with healthy patients as well is highly desirable. Devising such a method based on intensity myocardial profiles could be conducted by characterizing healthy and abnormal myocardium histograms. In previous works, healthy and scarred myocardial tissue distributions have been well described [10],[11],[12]. While a Rayleigh (or a Rician) distribution might appropriately model the normal tissue, hyper-enhanced infarcted areas are suitably modeled by a Gaussian one. Thus, the whole myocardium histogram consists on the resulting distribution obtained from the overlapped healthy and abnormal tissues. For the sake of simplicity, assumption of both distributions as Gaussian models have been extensively used [2],[13],[14],[15]. Hennemuth et al. [10] proposed the use of information criteria (Akaike and Bayesian ones) for histogram characterization. Thus, by assessing the best histogram model fitness normal or abnormal myocardiums could be distinguished. However, the main limitation of this approach regards expectation-maximization algorithm convergence. Due to the considerable distribution's overlap in diseased cases, the algorithm sometimes converge to a unique component model. Moreover, in small myocardial lesions, the scarred tissue distribution is obscured by the healthy one, turning the method inaccurate. These limitations suggest that the problem could be better address by extracting more complex features (and not only intensity-based ones). In this sense, some initial attempts using deep learning were performed in CT scans for detecting coronary artery stenoses [16].

### B. Myocardial Infarction Segmentation

Intensity based segmentation algorithms have been widely investigated and validated in clinical practice. In these techniques, the histogram thresholding is conducted in a semi-automatic [4],[7],[17] or automatic approach [14],[18]. Given that these methods cannot deal with the overlapping tissue distribution areas, several studies extended or combined them by using more sophisticated tools. Common works recombined the thresholding techniques [19],[20],[21] or used intensity features with connected component analysis [11],[15],[22], clustering [23],[24] or support-vector-machines (SVM) [25],[26]. Graph-cuts [12],[27], watershed [10],[28] and continuous max-flow [29],[30] algorithms have received researchers attention as well. Moreover, a deep learning based approach has recently been proposed [31].

Despite the vast techniques exploration, up to now there is no reference method for scar quantification [2] and just few of these techniques are used in clinical practice. The considered state-of-the-art (SOA) ones comprise the n-SD and FWHM, even when its variability, reproducibility and lack of expert agreement were highly discussed [8],[9]. For these reasons, the development of a robust technique able to accurately reproduce the experts delineations becomes highly valuable.

## III. MATERIAL AND METHODS

### A. Data Acquisition

One-hundred randomly chosen late-gadolinium enhanced MRI cases (20 healthy, 80 with attested MI) from the University Hospital of Dijon were included in this study. Gadolinium contrast solution (Dotarem, Guerbet, France) was administered to the patients between 8 to 10 minutes before conducting the study. Thirty-five percent of infarcted cases (n=28) presented micro-vascular obstruction areas. For all patients, a short-axis stack of cardiac images covering the whole left ventricle were acquired using one of the two clinical MRI devices with magnetic fields of 1.5T and 3T (Siemens Healthineers, Erlangen, Germany). A phase sensitive inversion recovery sequence with slice thickness of 8 mm and slice gap of 2 mm was performed. Voxel sizes differed among scans between 1.25 mm x 1.25 mm to 1.91 mm x 1.91 mm.

The dataset ground truths were delineated in each slice by an expert of the institution (AL) with more than 15 years of expertise in the field. The endocardial and epicardial boundaries were contoured (papillary muscles were included in the cardiac cavity as recommended [32]) and in pathological cases the scar tissue was annotated taking separate contours for enhanced and MVO areas. For assessing intra- and inter-observer annotations variability, infarcted areas of a random subset of pathological cases (50%, n = 40) were re-contoured by the same expert as well as by a second observer (TL, a cardiologist with 5 years of experience in the field).

### B. Proposed Method

In this work a new method for detection and quantification of myocardial infarction from short-axis cardiac LGE-MRI is presented. The method comprises two blocks which target the identification of diseased images and afterwards their segmentation. Firstly, healthy and pathological scans are discriminated using a classifier. Secondly, the scar tissue is segmented by an initial fast coarse segmentation followed by a voxel reclassification refinement strategy. The outputs of the algorithm are the delineated scarred areas with their corresponding clinical biomarkers. The whole method was implemented under Matlab® R2017b.

### 1. Data Pre-processing

Collected MRI scans present differences among them mainly in *i)* voxel size and *ii)* intensity values. While the former differences come from the setting of diverse scanning parameters, the latter differences may come from the use of different magnetic field devices (which account with diverse signal to noise levels) as well as by the inherent biological and anatomical patients variability. For homogenizing the scans all volumes were pre-processed by following three steps. Firstly, high-frequency noise was removed by using a spatial adaptive non-local means filter with automatic noise level estimation [33]. The chosen algorithm allows to tackle not only the intra-patient noise level differences in the scan, but also the inter-patient one observed by the use of different MRI magnetic field devices. Secondly, volumes were resliced to reach an homogenous voxel size of 1.25 mm x 1.25 mm x 8 mm (minimum voxel size found among patients). Thirdly, for reducing intra- and inter-patient intensity variability normalization within the epicardium inner region ([0-255]) was conducted. Since the contrast agent tissue concentration changes within time and the intensities become brighter from the mitral valve to the apex causing inter-slice variability [12], slice normalization was performed by taking into account the left ventricular myocardium and blood-pool regions. A subsequent enhancement by means of a gamma function was performed.

### 2. Myocardial Abnormality Detection

In this step a dichotomous classifier is built for discriminating healthy and pathological images. By using the epicardial mask and by assuming that the myocardial shape resembles a ring, the epicardium centroid is estimated. Afterwards, cropped images (size 89x89, 3-channel replicated) masked within the myocardium and centroid-centered are used as inputs of the classifier.

*Testing phase.* For achieving the classification task a three-step approach is conducted: i) Fine-tuned VGG19 [34] models are used for extracting informative features characterizing the myocardial images. ii) Extracted features followed a principal-component-analysis (PCA) dimensionality reduction by their projection into the learnt principal-components space. iii) Images are finally classify as healthy or infarcted by using SVM.

*Training phase I: Feature extraction.* The ImageNet pretrained VGG19 [34] model was chosen over other network architectures (such as VGG16, Resnet50, Resnet101, and GoogleNet) based on an exploratory performance analysis. In previous works, the model shows suitability and good adaptability for working in the medical domain [35],[36]. Since the main aim of this block is to devise a robust image classifier, only experiments with the achieved outperforming network are shown. The model was fine-tuned using MR images by preserving all layers and their corresponding weights with exception of the three ending fully-connected layers (FCL), whose neurons weights were re-learnt. Besides, after the last 1000-neuron FCL an extra 2-neuron FCL with a softmax layer was added for conducting the classification. The network training parameters are summarized in Table 1. For the replaced FCL, the learning rate was 30 times higher than the value shown on the table. Considering the dataset size limitations and with the aim of overfitting avoidance we *i)* performed data augmentation for increasing the training set by considering random geometric image transformations (rotation, shearing, flipping and scaling) *ii)* shuffled the training set in every epoch and *iii)* applied a random dropout [37] of 50% after each fully connected layer. Besides, for avoiding the classifier to produce biased class results, data imbalance was addressed by randomly sub-sampling the majority class until reaching the minority class size.

Once the network was fine-tuned, the whole training set was re-fed to the fitted network and image features were extracted from the 1000-FCL. Afterwards, the matrix of observations $X_{Nx1000}$ (where $N$ is the number of training samples) was built.

*Training phase II: Classifier fitting.* For conducting dimensionality reduction PCA was used by retaining the K principal components (K <<1000) that preserve 95% of the data variance. Afterwards, with the reduced observation matrix $X_{NxK}$ a SVM with linear kernel was fitted for classifying normal and infarcted myocardium images.

*Model Validation.* One-hundred random dataset splits were conducted in a class balanced 80-10-10% (training-validation-test) approach. For each training/validation set, fine tuning of VGG19, principal components decomposition and SVM fitting were conducted. Afterwards, over the test-set the label prediction was performed. Obtained classifiers were characterized and evaluated by means of a receiver-operating characteristic (ROC) curve analysis. Besides, to further validate whether the discriminant SVM rule could be randomly achieved, a one-hundred permutation analysis over *Healthy* vs *Diseased* cases was performed. Obtained ROC area under the curve (AUC) values were used as a global performance metric for comparing permuted and un-permuted classifier results.

TABLE I
SUMMARY OF THE CNN ARCHITECTURES USED IN THE FRAMEWORK AND THEIR CORRESPONDING PARAMETERS.

| Goal | NET | Patch | Loss Function | Optimizer | LR | M | Mini-batch | L2 | Epochs |
|------|-----|-------|---------------|-----------|-----|-----|-----------|-----|--------|
| Disease Detection | VGG19 [34] | 89x89 | Cross-entropy | SGDM | 1X10-4 | 0.9 | 16 | 1x10$^{-4}$ | 20 |
| Scar Segmentation | Zreik et al. [16]* | 49x49 | Cross-entropy | SGDM | 1X10-2 | 0.75 | 256 | 1x10$^{-4}$ | 50 |

LVM: Left Ventricular Myocardium; Net: Network Architecture; SGDM: Stochastic Gradient Descent with Momentum; LR: Learning Rate; M: Momentum; L2: L2 Regularizer; E: Epochs; * The elemental branch of the network was used instead of the whole architecture.

### 3. Myocardial Scar Quantification: Coarse Segmentation

Myocardial infarction was initially segmented after over-enhancing potential damaged regions using the non-parametric top-hat transform, as similarly conducted in other fields [38],[39],[40]. The enhancement was performed by using a sum of top-hats, which increases the contrast between dark and bright image areas (healthy and damaged regions respectively). The transform was applied in each slice using a 2D bar rotational structuring element with increasing variations of 30° and a constant length of 34 pixels. The enhancement reduced the overlapping areas of the healthy and scar distributions due to partial volume effect, helping the tissues discrimination by using Otsu's algorithm [18]. Structuring element shape, length and rotation-degree were empirically selected by maximizing the segmentation performance over the training set. Subsequently, a morphological opening (disk as structuring element, radius 1 pixel) was applied for removing small misclassified voxel clusters. The coarse segmentation workflow is shown on Fig. 1.

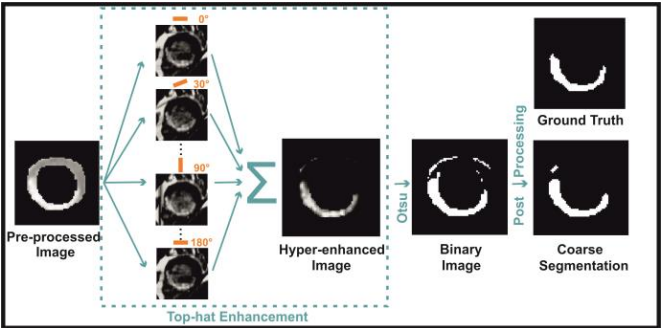

Fig. 1. Coarse segmentation workflow.

### 4. Myocardial Scar Quantification: Refined Segmentation

*Voxel reclassification phase.* After achieving an initial segmentation of the potential damaged areas, a voxel-level segmentation refinement was followed by using an ensemble of from-scratch trained CNNs. Although in the initial segmentation most lesions are detected within their core or more evident damaged areas, the method might provide misclassifications due to the overlapping healthy and infarcted intensity distributions. Thus, false positives (respectively negatives) removal (resp. inclusion) were tackled by a voxel reclassification approach using image patches centered at the voxels of interest. Voxels falling in a boundary region surrounding the coarse segmentation were re-classified with the aim of including in the analysis only the potentially misclassified ones. The boundary region was taken by subtracting to the morphologically dilated mask the morphologically eroded one. A disk as structuring element with radius of 2 pixels was considered. The degree of the morphological operations was experimentally chosen by assuring a mean sensitivity of 95% over the dilated masks on the training set. Voxel label prediction was achieved afterwards by majority voting after passing each patch over a seven-component CNN ensemble. The whole refinement segmentation workflow can be appreciated in Fig. 2.b).

*CNN's training phase.* The CNN architecture of Fig. 2.a) was used for voxel classification, which consists of a modified single-branch version of the architecture proposed for left ventricular myocardium segmentation in [16]. Unlike the original implementation, rectified linear units were used as activation functions [41]. For building the classifier, the network was trained from scratch by extracting boundary patches taken from the training set in a 50%-50% class balanced way. Ground-truth masks were dilated and eroded by using a disk of radius 5. Then, healthy-class patches were taken from the mask obtained after subtracting to the dilated mask the original ground-truth. Likewise, infarcted-class patches were extracted from the mask obtained after subtracting to the original infarction mask the eroded mask. In cases were the lesions were small (and hence the erosion operation degraded the whole mask) patches from the entire mask were taken. The reason for preferring boundary-close voxels instead of central ones relies on the difficulty for their detection, since partial volume effect and the so called gray-zone areas [15] make the tissue separation difficult. Voxel-centered patches were extracted with high information overlap (stride of 3 voxels) in order to help the network learning process [42]. A summary of the parameters used during the training phase is shown on Table I. In all cases, patches were zero-centered by subtracting the mean image of the training set. Overfitting avoidance and data balancing were conducted as described earlier in Section 3.2.1. The network training process converged after 50 epochs. The classifiers ensemble was built by training CNNs in a 7-fold cross-validation strategy over the considered training set. Networks were trained in the same fashion. Validation of the method was performed using 5-fold cross-validation (80-20% of patients as training/test sets respectively in each fold).

### 5. Myocardial Scar Quantification: MVO Inclusion

To include MVO areas, we took advantage of the pathological anatomy prior information provided by MVO structure. Indeed, MVO is represented by hypointense regions neighbouring the hyperintense areas [43]. Besides, infarction is always propagated from the endocardial cavity towards the epicardial one [3], assuring connectedness of the enhanced scar tissue volume with the blood-pool area. Mainly, MVO is found in the images as a dark cluster of voxels *i)* confined by the endocardial and enhanced areas or *ii)* fully enclosed in the enhanced region. For MVO inclusion, the union of the endocardial and hyperintense infarction masks was computed for finding all voxels clusters fulfilling the hypothesis. Afterwards, holes were filled providing a unique infarction segmentation mask including dark and bright pixel areas. The MVO inclusion strategy is illustrated in Fig. 2.c).

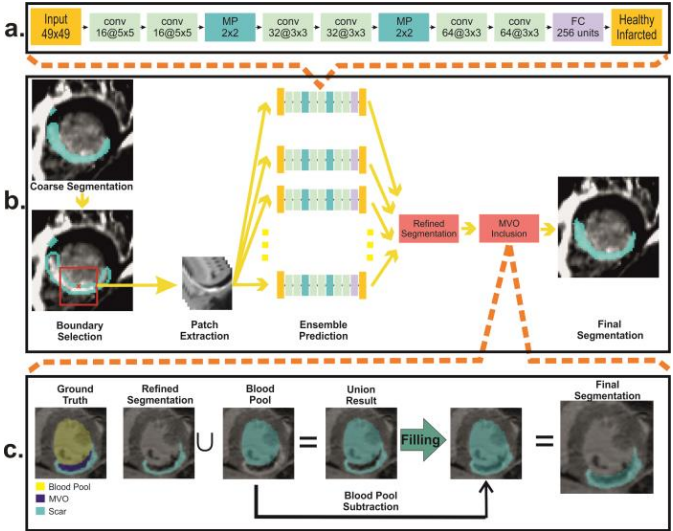

Fig. 2: Segmentation refinement block. a) CNN architecture used in the ensemble. b) Refinement segmentation workflow. c) Microvascular-obstruction inclusion workflow. conv: Convolutional Layer; MP: Max-Pooling Layer; FC: Fully Connected Layer.

#### 6. Comparison against State-Of-the-Art

In order to evaluate the proposed infarction segmentation algorithm performance, results were compared against nine standard algorithms widely used in clinical practice: the *n*-SD (n = 1, 2, ... , 6) [4], Otsu [18], FWHM (implementation of [22]) and Gaussian mixture models (with threshold at 2-SD above the mean healthy intensity [14]).

### C. Evaluation

Statistical analyses were conducted by first inspecting data behavior and then applying a t-Student or Mann-Whitney U tests when appropriate. For t-Student test conductions, normality was firstly checked by using the Shapiro-Wilk test while homoscedasticity was verified by data distribution inspection. Two-tailed tests with a 0.05 significance level were performed in all cases.

Image classification was assessed by sensitivity, specificity and accuracy metrics. Besides, characterization of the built classifier was evaluated by the area under the ROC curve. Mean and standard deviation of AUC values were reported. For assessing the model robustness in the ROC permutation analysis, the p - value was computed as follows:

$$p = \frac{\sum_{i=1}^{N} I(AUC_i^p, AUC_i^{np})}{N} \qquad (1)$$

where $N$ = 100 is the amount of data splits (and permutations) conducted, $AUC_i^p$ and $AUC_i^{np}$ are the obtained

AUC values for the permuted and un-permuted *i-th* dataset split respectively and the indicator function $I$ is defined as follows:

$$I(AUC_i^p, AUC_i^{np}) = \begin{cases} 1 \ if \ AUC_i^p \geq AUC_i^{np} \\ 0 \ if \ AUC_i^p < AUC_i^{np} \end{cases} \qquad (2)$$

Method's segmentation performances were assessed by Dice similarity indexes and 3D Haussdorf distances [32]. The scarred myocardial volume (cm$^3$) and percentage of infarcted myocardium (% $\frac{Vol_{scar}}{Vol_{myocardium}}$) were quantified for assessing clinical markers' estimation performances. Results were compared with the expert annotations by using Spearman correlation coefficient and Bland-Altman [44] analysis (mean and standard deviation of bias are provided).

## IV. RESULTS

### A. Myocardial Abnormality Detection

#### 1. Model Selection

The herein proposed classifier was chosen after comparing the model against a classical VGG19 fine-tuned one. In this latter model, the network was fine-tuned in the same fashion as described in Section 3.2.2 but the final prediction over the test-set was directly conducted by the network. Performance results for the models are shown in Table 2. The chosen model outperformed the VGG19 fine-tuned one in terms of sensitivity and accuracy in a maximum-a-posteriori prediction. Even more, the metric variances were lesser for the selected classifier.

#### 2. Classifier evaluation

The AUC value obtained on the ROC analysis was 0.95 ± 0.03 for the proposed classifier (Fig 3). The results obtained under 100-random splits scenarios show high performance stability and low variance. Since the classifier will be used to

TABLE II
MEAN (STANDARD-DEVIATION) PERFORMANCE METRICS OBTAINED FOR THE EXPLORED CLASSIFIERS UNDER THE 100-RANDOM SPLITS VALIDATION.

| Method | SE | Sp | Acc |
|---|---|---|---|
| Fine-tuned VGG19 | 84.41 (11.02) | **93.89 (6.79)** | 89.15 (5.36) |
| **Proposed** | **88.11 (6.54)** | 93.15 (4.84) | **90.63 (4.32)** |

Se: Sensitivity; Sp: Specificity; Acc: Accuracy.

decide whether or not the segmentation lesion search algorithm should be applied in each image, it is not equally important to have false positive or negative detections. Thus, each pathological image misclassified as a healthy one will not be assessed by the segmentation algorithm and their damaged areas will be lost from the analysis. On the other hand, misclassified healthy images into pathological ones might tend to produce an over-segmentation of the lesion. Under this scenario the classifier was set up for assuring high-sensitivity performances. When moving the decision rule threshold for addressing this goal we obtained for sensitivities of 90%, 92.5%, 95% and 97.5% corresponding specificity values of 90%, 85.4%, 73.3% and 57.3%.

The last experiment of this section involved a ROC permutation analysis. There were consistent AUC distribution differences between permuted and un-permuted data, which showed statistical significance ($p < 0.05$, paired t-Student test). For the random dataset configurations there was no permutation outperforming in AUC terms the original data

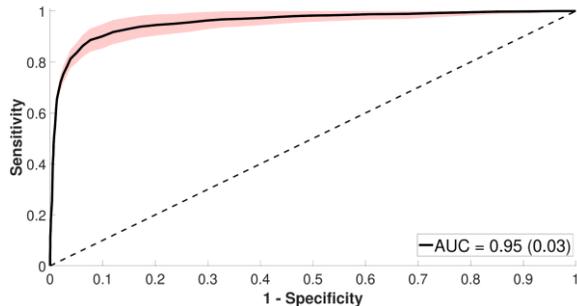

Fig. 3. ROC curve obtained after 100-random splits for the proposed classifier. The solid black line represents the mean AUC performance obtained, while the red area represents the variability AUC interval (mean SD). ROC: Receiver Operating Characteristic; AUC: Area Under the Curve.

configuration.

### B. Myocardial Scar Segmentation and Quantification

#### 1. Ensemble size selection

The idea of training an ensemble of classifiers using 7 CNNs is based on a comparative analysis conducted for different ensemble sizes. Results obtained for different ensemble models (with k = 1, 3, 5, 7, 9 components) are reported in Table 3. The coarse segmentation by itself achieved an overall Dice index of 73%. When the segmentation refinement was introduced, results improved until reaching a mean Dice index of 77.22% for the ensemble using 7 CNNs. It is noticeable that using an ensemble with more CNNs did not improve the segmentation performance. Besides, the ensemble of 7 CNNs performed closely in terms of Haussdorf distance ($< 1$ mm difference) to the best performing model (ensemble with 1 CNN). Consequently, after this experiment the number of CNNs was fixed to seven and from here on, all presented results are obtained under this chosen configuration.

#### 2. Segmentation Performance

Achieved segmentation performances are shown in Fig. 4. With the aim of comparing all the methods under similar working scenarios, segmentation metrics for all algorithms were computed over diseased images. Our algorithm obtained the highest Dice index when compared against the SOA method ones, achieving a Dice value of $77.22 \pm 14.3\%$ and considerably outperforming the best ranked SOA method (2-SD with Dice $70.49 \pm 16.48\%$). Besides, our proposal obtained the lowest Dice variance among all methods. Statistical significance was present in all Dice comparisons. When comparing performances in terms of Haussdorf distances, our method obtained $41.2 \pm 17.8$ mm (Fig 7B). The lowest Haussdorf values were obtained for the 4-SD and 5-SD methods ($30.8 \pm 16.9$ mm and $31.7 \pm 18.4$ mm respectively, $p < 0.05$). The achieved homoscedastic Haussdorf distance distributions showed similar variance levels for all the methods.

Qualitative segmentation results at different heart locations are shown in Fig. 5. As can be observed, the algorithm is robust for detecting the scar at different heart positions. Overall, less false-positives cluster of pixels were found for our method when comparing against the SOA ones. We can also notice the segmentation improvement obtained after the refinement step.

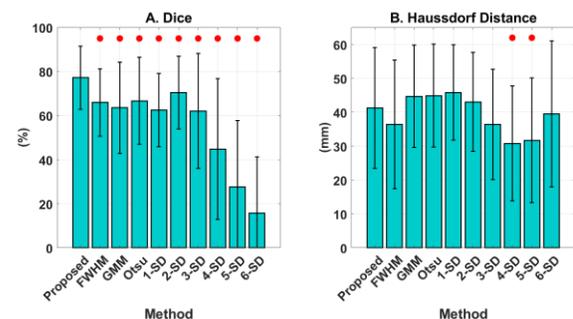

Fig. 4: Segmentation performances. FWHM: Full-width at half-maximum; GMM: Gaussian-mixture-model; n-SD: n-standard deviation thresholding from remote myocardium; * : $p < 0.05$ obtained by Mann-Whitney U-test.

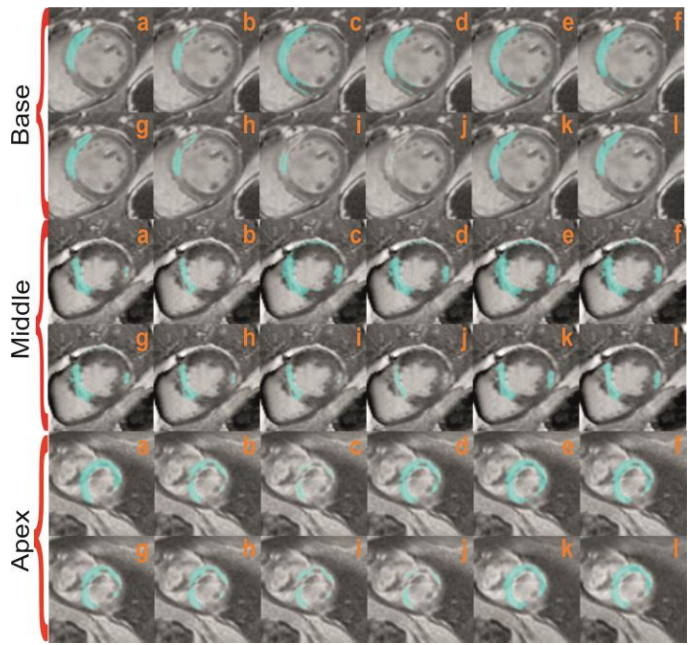

Fig. 5. Scar segmentations obtained per algorithm at different heart locations. a) Ground-truth. b) Full-width at half-maximum. c) Gaussian-mixture-models. d) Otsu. e) 1-SD f) 2-SD. g) 3-SD. h) 4-SD. i) 5-SD. j) 6-SD. k) Proposed coarse segmentation. l) Full proposed method.

TABLE III
MEAN (STANDARD-DEVIATION) SEGMENTATION PERFORMANCES OBTAINED
FOR THE COARSE SEGMENTATION FOLLOWED BY DIFFERENT ENSEMBLE SIZES.

| Method | Dice (%) | HD (mm) |
|---|---|---|
| Coarse | 73.0 (14.5) | 41.9 (17.3) |
| Coarse + Ensemble 1 | 76.3 (14.9) | **40.9 (18.0)** |
| Coarse + Ensemble 3 | 76.9 (14.7) | 41.4 (17.8) |
| Coarse + Ensemble 5 | 77.1 (14.4) | 41.3 (17.7) |
| Coarse + Ensemble 7 | **77.2 (14.3)** | 41.2(17.8) |
| Coarse + Ensemble 9 | 77.2 (14.3) | 41.3 (17.88) |

Ens: Ensemble size; Dice: Dice index; HD: Haussdorf distance.

Intra-observer variability for the infarcted volume and percentage of infarcted myocardium were -2.2 ± 7 cm³ (ρ= 0.98) and -1.0 ± 2.4 % (ρ= 0.973) respectively. On the other hand, the inter-observer variability for the infarcted volume and percentage of infarcted myocardium were 11.0 ± 7.04 cm³ (ρ= 0.915) and 5.2 ± 9.7 % (ρ= 0.9). The agreement of the different methods with the manual delineations in terms of clinical markers is summarized in Table 4. Estimation of the scarred myocardial volume as well as of the percentage of infarcted myocardium were consistently better for our proposal when compared against the SOA ones. For both considered metrics, our approach achieved the highest correlation values and lowest Bland-Altman biases. A relevant result is that our proposal was the only method in estimating the scar volume and percentage of infarcted myocardium by agreeing with the manual delineations. All the remaining methods obtained clinical markers estimations that statistically differed from the expert annotated ones.

### 3. Microvascular Obstruction Inclusion

In Table 5 the sensitivity of the different methods for detecting MVO areas are shown. Our proposal achieved the highest performance values and showed statistical significance when compared with all SOA methods with exception of the 1-SD one. A MVO segmentation example can be appreciated in Fig. 6, where our proposal's capability for the task is exposed. It can be highlighted the accurate segmentation of the hyper-enhanced area provided by the coarse pre-segmentation, with its improvement and MVO inclusion after the refinement approach. For the shown image, only our approach was able to deal with the no-reflow area.

## V. DISCUSSION

In this work, a new method for myocardial infarction detection and quantification in LGE-MRI is presented. Among the main novelties of our proposal we can point out: i) the generalization of the algorithm for working under healthy scenarios, ii) the automated detection of myocardial lesions and subsequent quantification, iii) the incorporation of a dedicated step for including MVO areas within the scar segmentation and iv) the validation of the algorithm on a large multi-field annotated LGE-MRI database. All these characteristics make our proposal a highly valuable tool with clinical transfer potential.

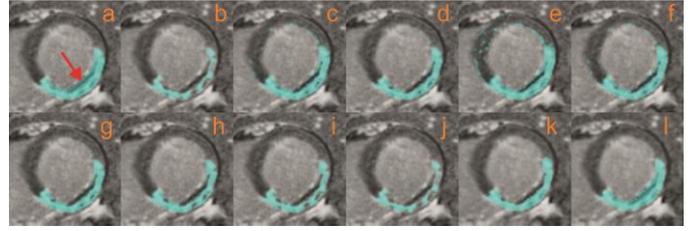

Fig. 6. Segmentation results for microvascular-obstructed areas per method. a) Ground-truth. b) Full-width at half-maximum. c) Gaussian-mixture-models. d) Otsu. e) 1-SD f) 2-SD. g) 3-SD. h) 4-SD. i) 5-SD. j) 6-SD. k) Proposed coarse segmentation. l) Full proposed method. The arrow indicates the microvascular-obstructed area.

With the aim of overcoming most current algorithm's limitation for working with healthy images, a dichotomic classifier for discriminating healthy and diseased cases was devised. The discriminant rule achieved high classification performance results. Our results suggest that features extracted with a CNN followed by a supervised classifier such as SVM, performs better than an end-to-end training of CNN models for classifying myocardial images. When the decision rule was assessed in terms of a ROC analysis, high AUC metrics with low variance were obtained, suggesting robustness of the proposed discriminant rule. Possible operative points providing high sensitivity were proposed, which will help in reducing the false positive lesions' detection in healthy images. Even more, results from the permutation analysis showed that the built classifier and the features used are informative for the addressed problem and cannot be achieved by a random chance configuration. All these findings support, consequently, the classifiers robustness as well as the method's reproducibility over different databases.

Segmentation of the infarcted masses was conducted in a two-step approach where the initial segmentation was later improved using deep learning. It is important to highlight the

TABLE IV
AGREEMENT BETWEEN METHODS AND THE MANUAL DELINEATIONS BY MEANS OF CLINICAL MARKERS.

| Method | SCARRED MYOCARDIUM VOLUME (CM3) | | | | % INFARCTED MYOCARDIAL VOLUME | | | |
|---|---|---|---|---|---|---|---|---|
| | Value | ρ | BA Bias | p-value | Value | ρ | BA Bias | p-value |
| Manual | 25.7 (19.4) | | | | 18.5 (12.7) | | | |
| FWHM | 17.6 (12.9) | 0.937 | -8.1 (8.7) | p < 0.001 | 12.7 (8.1) | 0.934 | -5.8 (6.0) | p < 0.001 |
| GMM | 32.7 (17.2) | 0.807 | 7.1 (14.6) | p < 0.001 | 24.2 (11.0) | 0.777 | 5.6 (9.5) | p < 0.001 |
| Otsu | 39.3 (20.8) | 0.907 | 13.6 (10.1) | p < 0.001 | 28.6 (12.1) | 0.885 | 10.1 (6.6) | p < 0.001 |
| 1-SD | 44.0 (24.4) | 0.932 | 18.3 (10.8) | p < 0.001 | 32.3 (15.8) | 0.923 | 13.8 (7.2) | p < 0.001 |
| 2-SD | 28.6 (18.7) | 0.9272 | 3.0 (9.4) | p < 0.01 | 21.1 (13.0) | 0.916 | 2.6 (6.4) | p < 0.001 |
| 3-SD | 18.4 (14.5) | 0.8749 | -7.2 (12.6) | p < 0.001 | 13.7 (10.4) | 0.847 | -4.9 (7.7) | p < 0.001 |
| 4-SD | 11.2 (11.1) | 0.7557 | -14.5 (15.3) | p < 0.001 | 8.4 (8.0) | 0.723 | -10.2 (9.4) | p < 0.001 |
| 5-SD | 6.4 (8.4) | 0.5672 | -19.3 (17.0) | p < 0.001 | 4.8 (6.0) | 0.565 | -13.7 ( 10.6) | p < 0.001 |
| 6-SD | 3.7(6.4) | 0.4581 | -21.9 (17.7) | p < 0.001 | 2.8 (4.4) | 0.471 | -15.8 ( 11.3) | p < 0.001 |
| **Proposed** | **26.6 (18.5)** | **0.945** | **1.0 (6.9)** | **0.196** | **19.1 (11.0)** | **0.945** | **0.5 (4.6)** | **0.314** |

Mean (standard deviation). FWHM: Full-width at half-maximum; GMM: Gaussian-mixture-model; n-SD: n-standard deviation thresholding from remote myocardium; BA: Bland-Altman; ρ: Spearman correlation coefficient; p - values obtained by a paired t-Student test.

TABLE V
MEAN (STANDARD-DEVIATION) SENSITIVITY FOR DETECTING MICROVASCULAR-OBSTRUCTION AREAS PER METHOD.

| FWHM | GMM | Otsu | 1-SD | 2-SD | 3-SD | 4-SD | 5-SD | 6-SD | Proposed |
|---|---|---|---|---|---|---|---|---|---|
| 18.8 (23) * | 54.6 (37.1) * | 57.7 (35.0) * | 63.6 (36.4) | 46.5 (38.9) * | 27.9 (34.0) * | 14.1 (25.1) * | 5.7 (14.6) * | 2.8 (9.1) * | **66.9 (40.5)** |

FWHM: Full-width at half-maximum; GMM: Gaussian-mixture-model; n-SD: n-standard deviation thresholding from remote myocardium. ∗ : p < 0.05 by means of Mann-Whitney U test.

novelty of this approach which was not only thought as a high-performance algorithm, but also as a modular transferable technique. Thus, high performance results were even obtained before conducting the segmentation refinement, achieving the coarse segmentation step a better agreement with the ground truth than the SOA methods. When the deep-learning based refinement was included, a consistent and statistical significant improvement in segmentation agreement was achieved. Even more, the low Dice variance showed homogeneity and adaptability of the method to different myocardial lesions configurations. When assessing the Haussdorf distance results high levels were obtained for most methods, which could be due to the 3D implementation of the metric. Nonetheless, the obtained distances were in the range reported by [30]. Our method obtained similar performances to most SOA algorithms excepting 4-SD and 5-SD, which achieved lower metrics (p < 0.05). These results are expectable since in these algorithms the intensity segmentation threshold is set very high. Thus, only highly hyperenhanced voxels belonging to the core necrotic tissue are detected and false detections coming from overlapped histogram areas are avoided. Given the fact that the Haussdorf distance is strongly affected by outliers [36], these methods may result benefited by this metric.

Promising results in terms of clinical markers were achieved with the proposed algorithm. The high correlation, low bias and the fact of being the only method agreeing in volumetric lesion quantification with the manual delineations suggest its appropriateness for working under clinical and medical conditions. It is important to highlight that our results shared the intra- and inter-observers variability ranges. On the other hand, supporting the findings of [8],[9] the SOA results showed very poor scar segmentation agreement with the manual delineations, characterized by low accuracies, high results variability and significant differences in volumetric tissue quantification.

Considering the novelty of the used dataset that contains no-reflow annotated cases, the inclusion of MVO areas within the infarcted segmented masks was compared between the different methods. Our approach was consistently superior for conducting this task, achieving the highest sensitivity performance and evidencing statistical significance when compared against the SOA approaches. The 1-SD method was the only exception, showing non-significant differences even when achieving lower performances. For this latter technique, the setting of a very low threshold for detecting myocardial scars favors MVO detection at the expenses of providing low overall performances.

With the aim of transferring the method to clinical scenarios, this algorithm is being implemented in the cardiac MRI post-processing software of CASIS company (https://www.casis.fr/). As a limitation of our proposal we can point out the two independent modules devised for lesion detection and quantification. Future goals will address the algorithm unification into a unique block while still preserving flexibility and modularity capabilities. Besides, for a fully automatic framework free of expert interaction the segmentation of the left-ventricle myocardium should be incorporated. In this work, given the complexity of the task and its high impact over the infarction biomarkers, manual contouring was preferred.

## VI. CONCLUSIONS

We propose a new method for infarction segmentation and quantification in LGE-MRI. The method overcomes several limitations of previous proposals from which the following points can be highlighted: *i)* repeatability, a limitation of semi-automatic approaches such as n-SD and FWHM methods, *ii)* detection of healthy and diseased slices, allowing to extend the method for working with healthy patients, *iii)* development of a novel and accurate technique for automatic delineation of the scar tissue, *iv)* incorporation of a no-reflow strategy for including these regions in the infarction quantification and *v)* validation on a large multi-field annotated database. The extensive statistical validation of the framework and its vast comparison against several current state-of-the-art methods turn this proposal into a robust and reliable tool with clinical transfer potential.

### ACKNOWLEDGMENTS

Ezequiel de la Rosa received an Erasmus+ scholarship from the Erasmus Mundus Joint Master Degree in Medical Imaging and Applications (MAIA), a programme funded by the Erasmus+ programme of the European Union.